\pdfoutput=1
\documentclass[11pt]{article}

\usepackage[]{acl}

\usepackage{times}
\usepackage{microtype}
\usepackage{subfigure}
\usepackage{graphics}

\usepackage{booktabs} % For formal tables
\usepackage{latexsym}
\usepackage{amsmath}
\usepackage{xspace}
\usepackage{amssymb}
\usepackage{graphicx}
\usepackage{multirow}
\usepackage{enumitem}
\usepackage{ulem}
\usepackage{float}
\usepackage{makecell}
\usepackage{tabularx}
\usepackage{stfloats}

\newcommand{\add}[1]{{\color{black}#1}}

\usepackage[T1]{fontenc}
\usepackage[utf8]{inputenc}

\usepackage{microtype}

\newcommand{\ie}{\textit{i.e.,}\xspace}
\newcommand{\eg}{\textit{e.g.,}\xspace}

\newcommand{\ce}{cross-entropy\xspace}

\newtheorem{finding}{Finding}

\title{Perplexity by PLM Is Unreliable for Evaluating Text Quality}

\author{Yequan Wang$^{1*}$, Jiawen Deng$^{2*}$, Aixin Sun$^{3}$, Xuying Meng$^{4}$\\
$^{1}$Beijing Academy of Artificial Intelligence, Beijing, China\\
$^{2}$CoAI Group, DCST, IAI, BNRIST, Tsinghua University, Beijing, China\\
$^{3}$School of Computer Science and Engineering, Nanyang Technological University, Singapore\\
$^{4}$Institute of Computing Technology, Chinese Academy of Sciences, Beijing, China\\
tshwangyequan@gmail.com, dengjw2021@mail.tsinghua.edu.cn,\\
axsun@ntu.edu.sg, mengxuying@ict.ac.cn
}

\begin{document}
\maketitle

\renewcommand{\thefootnote}{\fnsymbol{footnote}}
\footnotetext[1]{Indicates equal contribution}
\renewcommand{\thefootnote}{\arabic{footnote}}

\begin{abstract}
Recently, perplexity~(PPL) has been adopted to evaluate text quality, or more specifically fluency, of generated text in a few studies. 
A smaller PPL value means better text quality or better fluency. Through carefully designed experiments, we show that PPL is an unreliable measure for text quality. Specifically, we show that: (i) PPL of short text is more likely to be larger than that of long text.  (ii) Repeated text spans lead to lower PPL values although repeated spans often do not contribute to better text quality,  and (iii) PPL values can be largely affected by punctuation marks. Based on the findings, we further discuss the key issues in evaluating text quality using language models.
%However, we find that the PPL referee is unqualified and it cannot evaluate the generated text fairly for the following reasons:
%(i) The PPL of short text is larger than long text, which goes against common sense, (ii) The repeated text span could damage the performance of PPL, and (iii) The punctuation marks could affect the performance of PPL heavily. Experiments show that the PPL is unreliable for evaluating the quality of given text. 
%Last, we discuss the key problems with evaluating text quality using language models.
\end{abstract}

%===========================
\section{Introduction}
\label{sec:intro}
%===========================

The rapid development in natural language processing, particularly the success of pre-trained language models, has brought tremendous growth and progress to various text generation tasks. Examples include machine translation~\cite{DBLP:conf/acl/TuLLLL16,DBLP:journals/taslp/ZhangLSZX021}, question answering~\cite{DBLP:conf/emnlp/DuanTCZ17}, and generation-based dialog system~\cite{DBLP:conf/acl/TuLC0W022}. 
How to evaluate the quality of the generated text in a cost efficient manner has become a key challenge. 

%In this paper, we focus on automatic evaluation without user study or human rating. We consider task agnostic only in this paper.} 

Researchers have adopted various statistical metrics to evaluate the generated text. These measures include word-based measures like  BLEU~\cite{DBLP:conf/acl/PapineniRWZ02} and ROUGE~\cite{lin-2004-rouge}), character-based metrics like chrF~\cite{DBLP:conf/wmt/Popovic15}, and  embedding-based metrics like Vector Extrema~\cite{forgues2014bootstrapping} and Greedy Matching~\cite{DBLP:conf/bea/RusL12}. Specifically, 
BLEU reflects the ratio of overlapping $n$-grams to the total $n$-grams, denoting a precision-based measure. ROUGE and its variants, also evaluating text based on $n$-grams, are recall-based measures~\cite{DBLP:journals/csur/SaiMK23}.
Vector Extrema prioritizes informative words by taking the extreme value along each dimension. All these measures are widely adopted in many experiments and tasks. However, such statistical-based measures cannot well evaluate the creativeness, diversity, and complexity of texts, particularly in the scenario that the same semantic is expressed in different expressions, \eg different words/phrases, or different sentence structures. 

%\comment{I cannot understand this paragraph. It may need to cover these topics: (i) in the classical setting,  PPL is a measure to evaluate the quality of language models. We have training data to learn a language model, and then use the language model to compute PPL of test data. The smaller the PPL, then the better the language model. (ii)  why PPL is considered a reasonable measure? Then how PPL is computed when evaluating the generated text.  Given a piece of generated text, there is no ground truth text exist. It is to assume that pre-trained language model is pefect?}

In addition to the aforementioned statistical-based measures, perplexity (PPL) has also been used to evaluate the text quality or fluency in generation tasks. PPL is an intrinsic measure to quantify to what extent classical language models, \eg $n$-gram models, learn natural language~\cite{DBLP:conf/acl/MeisterC20}. Considering the large-scale pre-trained language models (PLMs) \eg BERT~\cite{DBLP:conf/naacl/DevlinCLT19} and GPT~\cite{radford2019language}, have well captured language knowledge, PPL has also been used to evaluate quality of generated text.\footnote{\url{https://huggingface.co/spaces/evaluate-metric/perplexity}} Given a PLM model and a sequence of generated text, perplexity reflects how likely the model is to generate this text sequence. If we assume a large PLM well captures language knowledge and is well-behaved, then the PPL value computed in this way could reflect the quality of the input sequence to some extent. 

%In addition to the aforementioned statistical-based measures, measures based on pre-trained language models (PLMs) are also being used~\cite{DBLP:journals/corr/abs-1907-12679}.  among which PPL is widely used to quantify how well language models learn natural language \cite{DBLP:conf/acl/MeisterC20}.
%It uses the cross-entropy of the probability distribution of predicted token and ground-truth token of text to evaluate the fluency of generated text. Here, the PLM could directly use mainstream models, such as BERT~\cite{DBLP:conf/naacl/DevlinCLT19}, GPT~\cite{radford2019language} \etal, so existing studies think that the PPL is able to reflect the fluency of generated text.

In this paper, we use PLM to compute PPL values of high quality sentences, as if these sentences were outputs from some generative models. Based on the distributions of PPL values, we claim that  PPL computed in this way cannot fairly evaluate  text quality. Specifically, we used GPT-2 model~\cite{radford2019language} to compute PPL  of sentences in WikiText-2 dataset.\footnote{\url{https://huggingface.co/datasets/wikitext}} As the sentences in WikiText dataset were extracted from verified good and featured articles on Wikipedia, we trust these sentences are of high quality. However, our experiments lead to the following findings. 
\paragraph{(i)} PPL is sensitive to text length, \ie PPL of short text is likely to be much larger than that of long text. On the other hand, the generated texts to be evaluated may have different lengths~\cite{meister-cotterell-2021-language}. Strictly speaking, text quality is independent of text length.   
\paragraph{(ii)} PPL is lower for text with repeated span(s). Generative text may contain repeated span(s). Although legitimate repeated text spans can be used to express emphasis in sentences, PPL cannot distinguish valid semantic emphasis in sentences from unreasonable straightforward repetitions.   
\paragraph{(iii)} PPL is sensitive to punctuation marks in sentences. Simply removing the last punctuation mark in a sentence may lead a significant increase in its PPL. On the other hand, removing the last punctuation from a sentence may only lead to a very small impact to human perception of the sentence.  

%For example, we have two texts, the former ends with punctuation, and the latter deletes the last punctuation. In theory, the qualified metric should compute the same or similar value. However, there is a significant difference between the PPL values of those two texts.

To the best of our knowledge, this is the first attempt to systematically analyze PPL for its suitability as a quality measure for generative text. % is not up to the task of text generation evaluation. And we detail the main reason \add{why PPL is unreliable} %for removing PPL is that PPL cannot handle the typical cases well. Last, we suggest a possible better evaluation direction.
%\add{Our paper summarizes the above vital findings of PPL failure. Further, we shed light on future potentially better evaluation metrics that should consider the following aspects to ensure fairness:}
Based on the findings, we call for more carefully designed metrics which are expected to be (i) not sensitive to length; (ii) sensitive to common mistakes, \eg unnecessary repeated text; (iii) not sensitive to minor punctuation changes. 
In other words, a measure of text fluency shall not be much affected by text length, while penalizing unnecessary text spans and not attending to non-significant punctuation marks.

%===========================
\section{Preliminary and Experiment Setup}
%===========================

%In this section, we briefly introduce the method used to evaluate the performance of PPL. 
In our experiments, we follow the mainstream approach  using GPT-2~\cite{radford2019language} as the pre-trained language model to calculate PPL. More specifically, we use the GPT2-large model.

%Given an input sentence $[w_1, w_2, \dots, w_n]$, we 

Given an input sentence $s$, we get its token sequence $s =[t_1, t_2, \dots, t_m]$ with $m$ size by the PLM. We use \add{GPT2-large} to compute the PPL:
\begin{equation}
    \mathcal{P}_1, \mathcal{P}_2, \dots, \mathcal{P}_m = \text{GPT-2}([t_1, t_2, \dots, t_m]),
    \label{eqn:gpt-2}
\end{equation}
where $P_i$ denotes the predicted probability of the $i$-th token. PPL of the input sentence $s$ is computed as the cross entropy of each token:
\begin{equation}
    \text{PPL}(s) = \exp\left \{
        \frac{1}{m}\sum_{i=1}^m \text{\ce} (t_i, \mathcal{P}_i)\right \}.
    \label{eqn:ce}
\end{equation}
Here, $ \text{PPL}(s) $ is the perplexity value of the input sentence $s$.

The sentences in our experiments are from the test split of the \textit{WikiText-2 dataset}. We filter sentences with fewer than 3 words to avoid extremely short sentences. As the result, we have  2,786 texts left in our experiments. The maximum, minimum, and average lengths are $481$, $3$, and $86.52$ tokens, respectively.

As sentences in WikiText were orginally from a set of carefully selected high quality Wikipedia articles, we assume the quality of all these sentences is high. Accordingly, if PPL computed by PLM is a suitable text quality measure,  we expect a reasonably stable PPL value for all these sentences. On the other hand, if the PPL values of these high quality sentences are spread in a large range, then PPL values may not well reflect the text quality of generative text.  

%We now evaluate the performance of PPL to prove that the PPL is not qualified for generated text evaluation.
%\add{In the experiments, we use the testing data from the test split of \textit{WikiText-2 dataset}  and take \textit{GPT2-large} as the PLM.}

%As we mentioned in Section~\ref{sec:intro}, PPL performs \add{unreliably} %badly 
%for mainly three reasons, including (i) text length, (ii) repeated text span, and (iii) punctuation mark. Next, we  experiment on each scene.

%===========================
\section{Experiments and Findings}
%===========================

%===========================
\subsection{PPL vs Text Length}
\begin{figure}
  \centering
\includegraphics[width=0.95\linewidth]{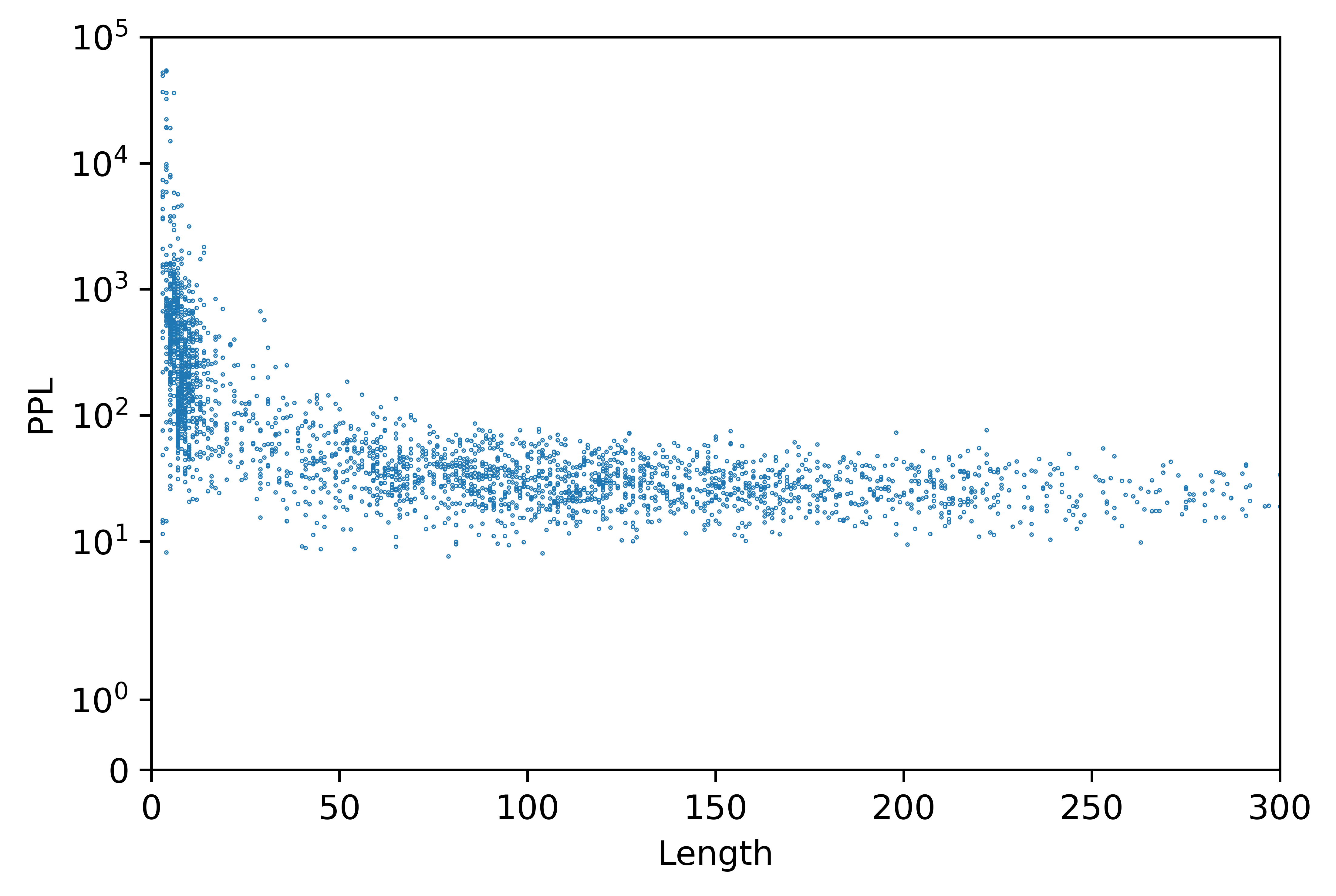}
  \caption{The PPL of text with different lengths. The $x$-axis denotes text length in number of tokens, and the $y$-axis is PPL value in log scale.}
  \label{fig:text_length}
\end{figure}

%===========================
We first evaluate whether PPL is sensitive to text length of high quality text. In human perception, text quality is not strongly correlated with text length. Given the high quality sentences from WikiText, we expect a stable PPL value for all sentences. Figure~\ref{fig:text_length} plots the PPL values of text against their lengths in number of tokens. Note that the PPL values on $y$-axis are in log scale.

%To evaluate the performance of PPL when meeting texts of different lengths, we design this simulation.

%Given sentence $s=[t_1, t_2, \dots, t_m]$, we use Equation~\ref{eqn:gpt-2} and \ref{eqn:ce} to calculate the PPL value. In human cognition, in general, the length of text has little relationship with the quality of the generated text. In other words, the PPL value should remain stable and not change with the length of the text.

%We know that text quality is not very related to text length. If the evaluation metric is sensitive to text length, then the value of this metric is not convincing. So we conduct the experiment on text length to find the inconsequence of PPL. Figure~\ref{fig:text_length} details the changing trend of PPL as the text length changes.

\begin{finding}
PPL values are unstable for short texts, and become lower along the increase of text length.
\end{finding}

Observe that, a good number of sentences are short sentences, shorter than 25 tokens. These short sentences have a very wide range of PPL values with the majority in the $10^2 - 10^3$ range. Then along the increase of text length, the PPL values become lower. Nevertheless, fluctuation of PPL values exist for sentences in similar length, although the 
amplitude is not as large as when the text is short. 

If we consider text quality or fluency does not have high correlation with text length, then the PPL value distribution of these high quality sentences are unexpected. In this sense, we claim that PPL is not a good measure for text quality of generated text. In text generation tasks, the generated sentences may be of very different lengths. The changes in PPL values of generated sentences could be due to the text quality as well as simply the lengths of the sentences. 

%Obviously, it does not fit our common sense. 
%\add{Even, when the length of the text becomes longer, the text is more difficult to understand, resulting in higher PPL.}
%However, the experimental result reveals that the PPL has an opposite trend when meeting the text with different lengths. More importantly, the length of the generated text is variable so the problem is more serious.

\begin{table}
\centering
\caption{The PPL of repeating the last $q$ tokens for $k$ times. Increase rate (\%) denotes the proportion of the increase in PPL with repeated spans against the original. \textit{Normal Ratio} denotes \textit{ratio of samples with increasing ppl value} after the repeating operation.}
\scalebox{0.80}{
\begin{tabularx}{.58\textwidth}{
    m{.02\textwidth}<{\raggedleft}
    m{.02\textwidth}<{\raggedleft}|
    m{.08\textwidth}<{\raggedleft}
    m{.08\textwidth}<{\raggedleft}
    m{.08\textwidth}<{\raggedleft}
    m{.15\textwidth}<{\centering}
    % m{.08\textwidth}<{\raggedleft}
    }
\toprule
\textbf{$q$}  & \textbf{$k$}  & \textbf{PPL\_avg} & \textbf{PPL\_std} & \textbf{Len\_avg} & \textbf{Normal\_ratio}\\\midrule
0  & 0  & 411.99   & 3546.23  & 86.52   &   --     \\
\midrule
1  & 1  & 235.39   & 1333.33  & 87.52   & 70.17   \\
1  & 3  & 87.78    & 250.82   & 89.52   & 24.59   \\
1  & 9  & 36.48    & 363.83   & 95.52   & 0.54    \\
1  & 12 & 34.93    & 583.00   & 98.52   & 0.36    \\
% 1  & 15 & 33.64    & 690.70   & 101.52  & 0.22\%    \\
\midrule
5  & 1  & 50.21    & 70.96    & 91.48   & 39.16   \\
5  & 3  & 21.85    & 13.09    & 101.41  & 0.86    \\
5  & 9  & 10.37    & 6.97     & 131.18  & 0.0     \\
5  & 12 & 8.35     & 5.75     & 146.07  & 0.0     \\
% 5  & 15 & 7.01     & 4.87     & 160.96  & 0.0\%     \\
\midrule
10 & 1  & 35.74    & 59.70    & 95.61   & 2.40    \\
10 & 3  & 14.43    & 8.22     & 113.79  & 0.0     \\
10 & 9  & 6.13     & 4.04     & 168.33  & 0.0     \\
10 & 12 & 4.84     & 3.15     & 195.60  & 0.0     \\
\bottomrule
\end{tabularx}}

\label{tab:repeat_last_tokens}
\end{table}

%===========================
\subsection{PPL vs Repeated Text Span}
%===========================

%Does PPL make sense when the text lengths are at the same level? 
It is our understanding that if a generative model does not behave well, the model may cause duplication in the generated text, particularly the later part of a sentence. For example, to answer ``\textit{Can you tell me what natural language processing is?}'', a dialog system may generate ``\textit{Natural language processing is a sub-field of linguistics, computer science science science...}''. The repeated text span ``science science'' damages the semantics of the text. A qualified referee would have the ability to detect such unreasonable repetition, and give a low quality rating for such case.

To mimic such common mistakes in generative models, we design the following experiments. Given an original sentence $s=[t_1, t_2, \dots, t_m]$ with $m$ tokens, we repeat the last $q$ tokens for $k$ times to generate a new sentence $s_1$. Accordingly, sentence $s_1$ has $ m + q \times k $ tokens.

%where the last $N$ tokens of the text are repeated $K$ times.

%To evaluate the performance of PPL, we design the following rule to imitate this kind of situation.

%Given , we repeat the last $q$ tokens to generate the sentence $s_1$:
%\begin{equation}
%    s_1 = [s, s_l\otimes e_k],
%\end{equation}
%where $ s_l = [t_{m-q+1}, \dots, t_{m}]$ denotes the latter text with $ q $ tokens.
%The operation $\otimes$ here means: $s_l\otimes e_k=[s_l, s_l, \dots, s_l]$, that is, the operation repeatedly concatenates $s_l$ for $k$ times, where $e_k$ is a vector with $k$ $1$s~\cite{DBLP:conf/emnlp/WangHZZ16}. Then we could use Equation~\ref{eqn:gpt-2} and \ref{eqn:ce} to compute the PPL of new sentence $s_1$ with $ m + q * k $ size.

%A big problem for the generation model is the unusual repetition of generated text. 

\begin{figure}
  \centering
  \includegraphics[width=0.95\linewidth]{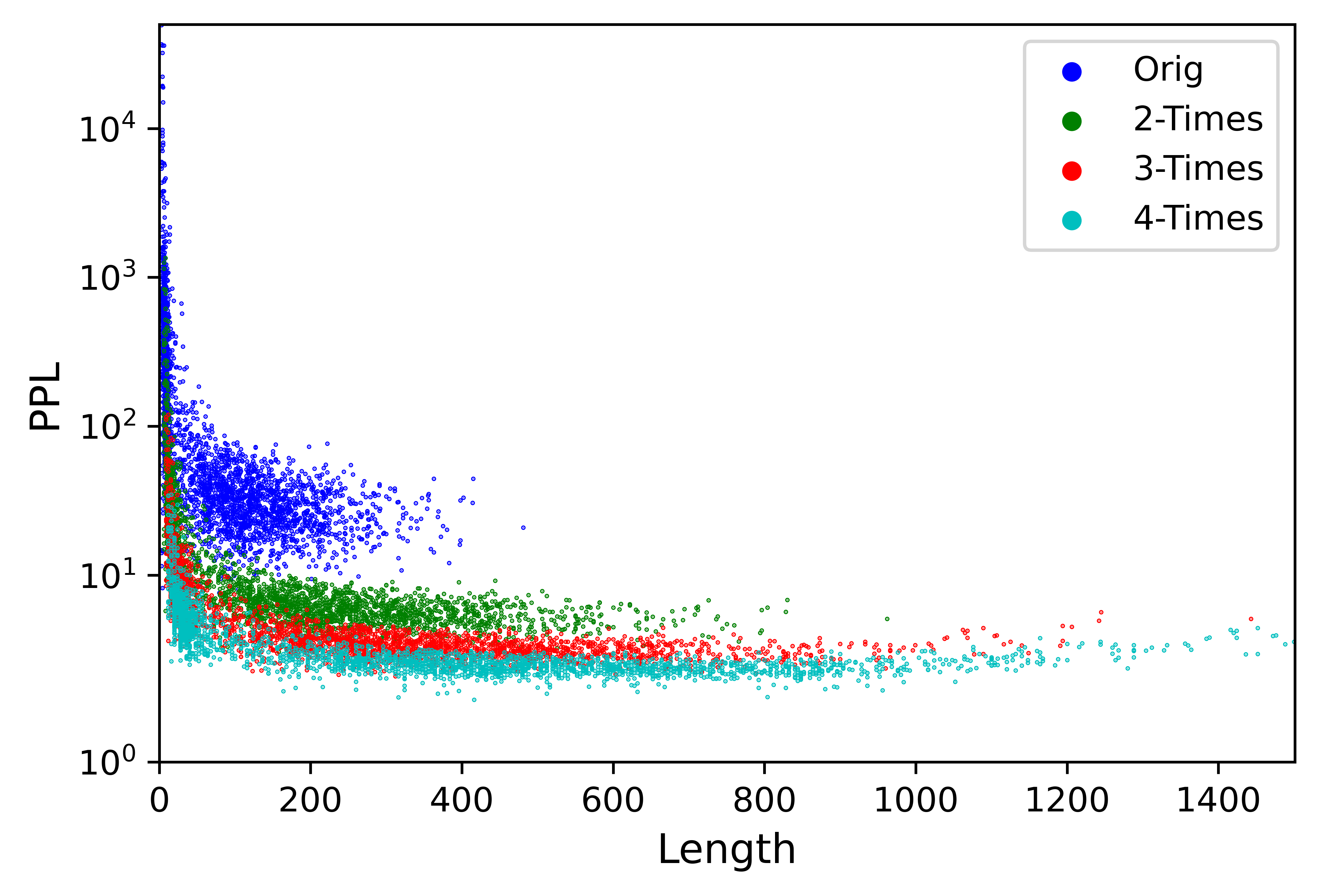}
  \caption{The PPL of the original text and repeating the entire text $2$, $3$, and $4$ times.}
  \label{fig:text_repeated_time}
\end{figure}

Table~\ref{tab:repeat_last_tokens} reported the averaged PPL (PPL\_avg) values of the original sentences in the collection ($q=0$, $k=0$), and the averaged PPL of the sentences with different repetitive spans. The table also reports the standard deviation of the PPL values, the average length of the original sentences and the sentences with repetitive spans. To assess the proportion of PPL that behaves normally, we define \textit{Ratio of samples with increasing ppl value} shown as \textbf{Normal\_ratio} in Table~\ref{tab:repeat_last_tokens}.
Figure~\ref{fig:text_repeated_time} plots the distribution of PPL values of the original text, and that of repeating the entire text 2, 3, and 4 times.

%gives the experiment result of repeating the entire text 2, 3, and 4 times. 

%Both Figure~\ref{fig:text_repeated_time} and Table~\ref{tab:repeat_last_tokens} tell us that 

\begin{finding}
The PPL value gets lower with more repetitive tokens.
\end{finding}

Table~\ref{tab:repeat_last_tokens} shows that the PPL values become lower when more repetitive tokens are included in the text. The reduction in PPL values are significant, even when only the last token is repeated for one time. Figure~~\ref{fig:text_repeated_time} shows that simply repeating the entire sentence multiple times leads to significant lower PPL values.

Meaningless repetition does not contribution positively to text quality, and may damage a sentence structure, making it grammatically incorrect. If PPL were a perfect measure for text quality, we expect PPL values of the sentences with repetitive spans to be higher than the original high quality text.  Our experiments show the opposite. More importantly, the normal ratio shows significant decreases when the repeated times increase. This phenomenon reveals that perplexity is unreliable.

%This experiment proves that the PPL could not be competent when meeting the unusual repeated text.

Note that, text repetition could be used to express emotions such as emphasis, anger, or other purposes  of highlights. However, in our experiment design, given the large number of sentences in the datasets, simply repeating the last few tokens in every sentence certainly is not in this category of language use.  In other words, the repeats in our experiment design is largely meaningless repeats and shall adversely affect text quality. We also note that, by repeating the last few tokens, the sentences become longer, which would lead to lower PPL based on our first finding. 

%When meeting this kind of text, we do not want the metric changes a lot. However, the experimental result tells us that \add{PPL value always drops sharply in case of replication}. This is because the used PLM model (\ie GPT-2) is trained by lots of language text, and such special text representations are relatively rare. 

%===========================
\subsection{ PPL vs Punctuation Mark}
%===========================

\begin{table}
\centering
\caption{The effect of punctuation marks. w/o All: Removing all punctuation marks; w/o Last: Removing the last punctuation mark.}
\scalebox{0.80}{
\begin{tabularx}{.58\textwidth}{
    m{.087\textwidth}<{\raggedright}
    m{.08\textwidth}<{\raggedleft}
    m{.08\textwidth}<{\raggedleft}
    m{.08\textwidth}<{\raggedleft}
    m{.08\textwidth}<{\centering}
    }
\toprule
Data        & \textbf{PPL\_avg} & \textbf{PPL\_std} & \textbf{Len\_avg} & \textbf{Normal\_ratio}\\\midrule
Orig.       & 169.3     & 1009.3    & 102.0     & --
\\
w/o Last    & 279.5     & 2195.5    & 101.1     & 12.51
\\
w/o All     & 690.2     & 4706.9    & 88.1      & 13.70
\\
\bottomrule
\end{tabularx}}
\label{tab:remove_all_punctuation}
\end{table}

%Is PPL sensitive to punctuation? 
We know that punctuation marks are important for human reading and understanding. We design two sets of experiments to evaluate the impact of removing punctuation marks in sentences to PPL. In the first set of experiments, we remove all punctuation marks from a sentence $s$, and the resultant sentence is $s_2$. In the second set of experiments, we remove the last punctuation mark from the original sentence $s$, and we get sentence $s_2$.\footnote{We observe that very few sentences ends with two continuous punctuation marks. For such sentences, only the last punctuation mark is removed, to be consistent with all other sentences.} 

In our understanding, removal of all punctuation marks from a sentence would make the sentence difficult to read if the original sentence contains many punctuation marks. However, removing the last punctuation typically does not significantly affect reading or its semantic meaning, for most sentences. In this experiment design, the first set of experiments serves as a reference. Our main focus is on the PPL values changes in the second set of experiments.

\begin{figure}
  \centering
  \includegraphics[width=0.95\linewidth]{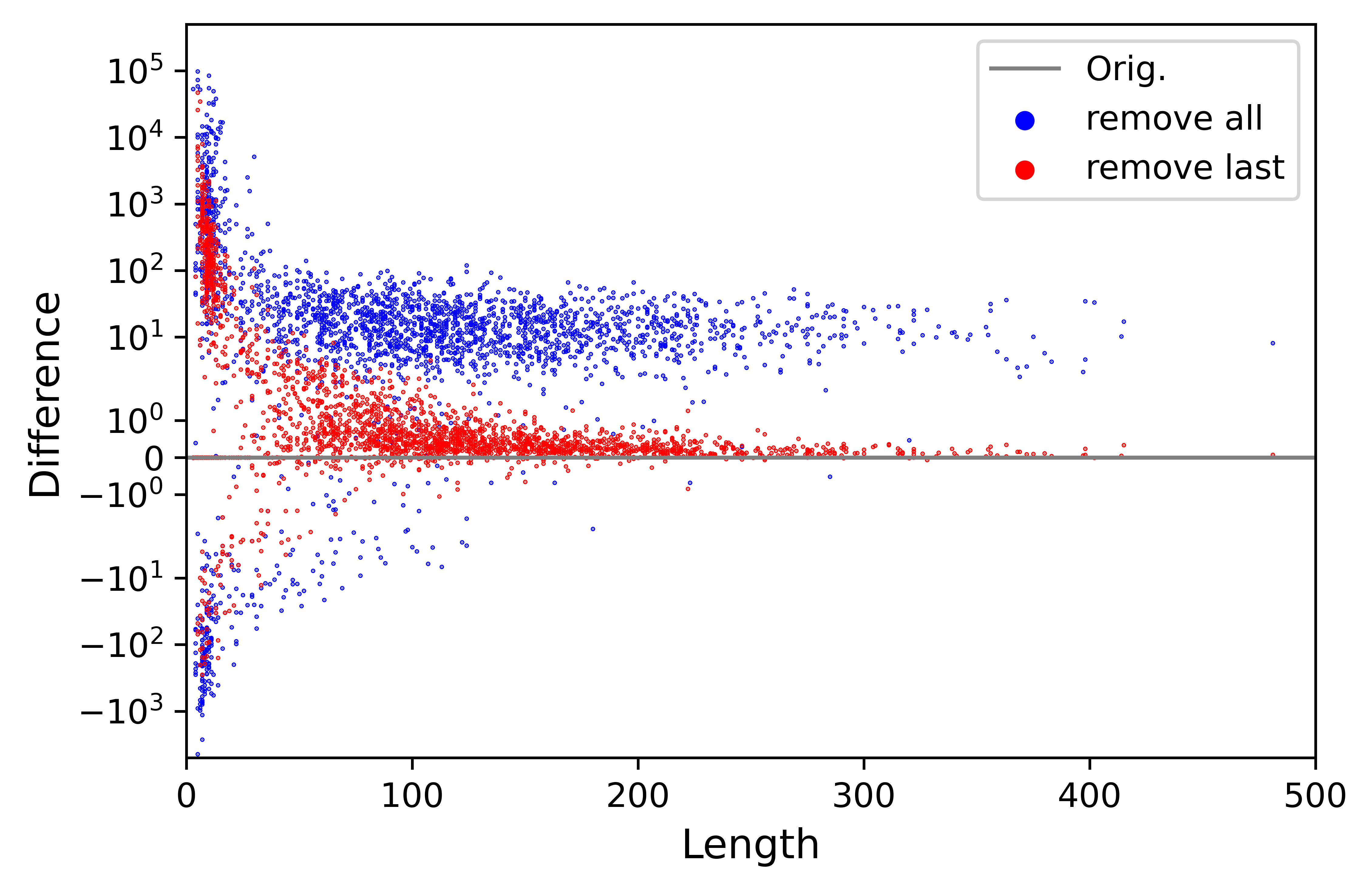}
  \caption{PPL changes due to removal of punctuation  marks against the original sentence. Difference in PPL = PPL of sentence after punctuation removal - PPL of original sentence. $s$-axis is the length of the original sentence.}
  \label{fig:text_remove_mark}
\end{figure}

Table~\ref{tab:remove_all_punctuation} shows the averaged PPL of the original sentences, and also the sentences after removing punctuation marks. PPL values shown in Table~\ref{tab:repeat_last_tokens} represent the results of length-filtered samples, and this table shows no filtering, so the original PPL values in Table~\ref{tab:remove_all_punctuation} are different from Table~\ref{tab:repeat_last_tokens}.
Figure~\ref{fig:text_remove_mark} plots the differences of PPL values with reference to the original sentences, due to punctuation marks removal.

\begin{finding}
Removing the last punctuation mark usually does not significantly affect the text quality, but results in significant changes in PPL values.
\end{finding}

%Table~\ref{tab:remove_all_punctuation} shows that over $13\%$ sentences have an unreasonable tendency when deleting all punctuation marks.
%Unfortunately and importantly, both Table~\ref{tab:remove_all_punctuation} and  Figure~\ref{fig:text_remove_mark} show that if we add or remove a punctuation mark that does not affect the semantics of language, the PPL value has a significant change. This is what we don't want to see. But it happens in the current PPL. 

As discussed, we would expect a minor or negligible change in PPL values if only the last punctuation mark is removed. Table~\ref{tab:remove_all_punctuation} and Figure~\ref{fig:text_remove_mark} tell us differently. That is, PPL is very sensitive the punctuation marks. Removal of the last punctuation marks does not lead to significant change to the sentence meaning or sentence length. The PPL changes significantly, particularly for short sentences.

%===========================
\section{Discussion}
%===========================
Through experiments of PPL values on high quality text, we show that PPL value is sensitive to text length. The good news is that PPL tends to be relatively stable if  text length is greater than $100$. Unfortunately, text length is not the only factor that is independent of text quality. Our experiments also show that PPL is unexpected to favor sentences containing meaningless repetitive text spans, and is sensitive to punctuation marks.

%In the above experiments, we first study the effect of text length. The results tell us that the PPL value is sensitive to the text length. Obviously, this problem damages the fairness of PPL. Meanwhile, the good news is the PPL tends to be stable if the text length is greater than $100$. Then, we conduct the next experiment to study the effect of unusual repeated text span. 
%However, PPL not only fails to remain stable, but also deviates negatively from the desired direction. Lastly, we study the effect of punctuation marks. Experimental results show that PPL still cannot achieve the desired effect. To give an intuitive description, the case study shown in Section~\ref{sec:app:case_study} is conducted.

All these results strongly suggest that PPL by PLM is unreliable for evaluating text quality.
We argue that a fair metric should well consider the impacts of text length, meaningless text repetition, and punctuation perturbation, among the other factors that may contribute to text quality.
In particular, diversity is an important indicator of text quality. The fluctuations of PPL values due to text repetition could be mitigated with text diversity, \eg penalizing the scoring based on the frequency of n-gram occurrences, and even penalizing consecutive occurrences of the same expression, if such a measure is well designed.
In short, we believe it is more reasonable to evaluate text quality with a collection of measures which check and balance each other. The widely used precision and recall in Information Retrieval is an example.  
%it is very important that a new metric to consider both PPL and diversity in a way like F1 score.
Lastly, for text quality, the correct use of punctuation is undoubtedly essential. But in the case of lower text quality, a good evaluation metric should focus more on the word coherence instead of the non-significant punctuation \eg the full stop at the end of a sentence. %that will not interfere with one's comprehension.}
%Considering these factors, a reliable evaluation metric is expected to be developed to assess the quality of a text.

%===========================
\section{Conclusion}
\label{sec:conclusion}
%===========================

In this paper, we evaluate PPL by PLM as an evaluation of text quality. Through experiments, we find that PPL by PLM is sensitive to text length, cannot well handle the common problem of unusual text duplication, and could be easily affected by punctuation marks. To this end, we consider PPL by PLM is unreliable for evaluating quality of text by generative models. We suggest a few factors for consideration when developing a new evaluation metric for text quality.

%===========================
\section*{Limitations}
%===========================

%In this paper, we \add{present} %address 
%the typical scenarios that PPL could not perform fairly when evaluating the quality of generated text. That is why we need to carefully consider whether to use this metric. 
%Further, we summarize the vital problem of PPL failure, then give suggestions to improve the validity and fairness of the evaluation.

There are two main limitations in this work. First, the three factors \ie text length, repetitive spans, and punctuation marks, are from a full coverage of a comprehensive understanding of text quality. Nevertheless, these are the factors that can be controlled through experiments, to study the behavior of PPL by PLMs. Second, we are not able to provide a more meaningful alternative measure for text quality. 

%However, the limitation is that we do not give a detailed solution. And in the future, we will continue to study it.

%===========================
\section*{Acknowledgments}
%===========================

This work was supported by the National Key R\&D Program of China (2020AAA0105200) and the National Science Foundation of China (NSFC No. 62106249, 61902382, 61972381).

\bibliography{anthology,custom}

\clearpage
\appendix

\section{Appendix}
\subsection{Case Study}
\label{sec:app:case_study}
\label{sec:appendix}

\begin{table*}[t]
\centering
\caption{Cases of unfair PPL when meeting texts with varying lengths. \textit{Combine} denotes the original text from the \textit{Wikitext-2 dataset}. \textit{Split1} and \textit{Split2} are the sub-sentences obtained by the slicing operation. }
\scalebox{0.72}{
\begin{tabularx}{1.35\textwidth}{
    m{1.1\textwidth}
    m{.1\textwidth}<{\centering}
    m{.05\textwidth}<{\raggedleft}}

\toprule
\textbf{Text}   & \textbf{Len.}   & \textbf{PPL}  
\\\midrule
Split1:\quad \textcolor{blue}{Over the course of his reign , Nero often made rulings that pleased the lower class .}
& 17    & 77.3  \\
Split2:\quad \textcolor[rgb]{0.502,0,0.502}{Nero was criticized as being obsessed with personal popularity .}
& 10    & 187.9 \\
Combine:\quad \textcolor{blue}{Over the course ... the lower class . }
\textcolor[rgb]{0.502,0,0.502}{Nero was criticized ... with personal popularity .}                 
& 27    & 60.6  \\\midrule

Split1:\quad \textcolor{blue}{Lesnar appears in the video games WWE SmackDown ! Shut Your Mouth , WWE SmackDown !}
& 16    & 26.3  \\
Split2:\quad \textcolor[rgb]{0.502,0,0.502}{Here Comes the Pain , Madden NFL 06 , UFC 2009 Undisputed , UFC Undisputed 2010 , WWE ' 12 , WWE ' 13 , WWE 2K14 , WWE 2K15 , WWE 2K16 , and WWE 2K17 .}
& 38    & 10.4  \\
Combine:\quad \textcolor{blue}{Lesnar appears in ... , WWE SmackDown ! }
\textcolor[rgb]{0.502,0,0.502}{Here Comes the Pain , ... and WWE 2K17 .}
& 54    & 8.7   \\\midrule

Split1:\quad \textcolor{blue}{The Great Fire of Rome erupted on the night of 18 July to 19 July 64 .}
& 17    & 62.8  \\
Split2:\quad \textcolor[rgb]{0.502,0,0.502}{The fire started at the southeastern end of the Circus Maximus in shops selling flammable goods .}                                                                     
& 17    & 65.7  \\
Combine:\quad \textcolor{blue}{The Great Fire ... 19 July 64 .} 
\textcolor[rgb]{0.502,0,0.502}{The fire started ... selling flammable goods .}
& 34    & 31.9  \\\midrule

Split1:\quad \textcolor{blue}{This isn 't acceptable to get to where we want to go . But what does that really mean ?}
& 20    & 40.9  \\
Split2:\quad \textcolor[rgb]{0.502,0,0.502}{It 's not just get better defensively , it is , if we give up 3 less baskets a game , then we will be at 40 percent field goal percentage defense which will be top 20 in the country}
& 40    & 44.3  \\
Combine:\quad \textcolor{blue}{This isn 't acceptable ... that really mean ? }
\textcolor[rgb]{0.502,0,0.502}{It 's not just ... in the country}
& 60    & 32.0  \\\bottomrule

\end{tabularx}}
\label{tab:example_unfairlen}
\end{table*}

\begin{table*}[t]% [tbph]
\centering
\caption{Cases for unreliable PPL. "$q , k $" denotes repeat the last $q$ tokens $k$ times. "\textit{2-Times}" denotes repeat the whole text. "\textit{w/o Punc.}" 
and "\textit{w/o Last Punc.}" denote removing all punctuations and the last punctuation, respectively.}

\scalebox{0.72}{
\begin{tabularx}{1.35\textwidth}{
    m{.03\textwidth}<{\raggedright}
    m{.52\textwidth}
    m{.03\textwidth}<{\centering}
    m{.06\textwidth}<{\raggedleft}
    m{.06\textwidth}<{\raggedleft}
    m{.06\textwidth}<{\raggedleft}
    m{.06\textwidth}<{\raggedleft}
    m{.09\textwidth}<{\raggedleft}
    m{.07\textwidth}<{\raggedleft}
    m{.10\textwidth}<{\raggedleft}
    }
    
\toprule
\textbf{ID}
& \textbf{Text} & \textbf{Len.} & \textbf{Orig.} 
& \textbf{$q1, k3$} & \textbf{$q1, k9$} 
& \textbf{$q5, k3$} & \textbf{2-Times} 
& \textbf{w/o Punc.} & \textbf{w/o Last Punc.}  
\\\midrule
1 & 
To use absolutely no word that does not contribute to the presentation .                  
& 13    & 140.1 & 64.8  & 24.0  & 20.5  & 17.0  & 106.6 & 106.6
\\\midrule
2 & Coastal service and riverine vessels , including ' floating batteries ' and ' monitors' 
& 14    & 321.4 & 255.1 & 41.2  & 53.6  & 31.1  & 282.3 & 235.0
\\\midrule
3 & = = Plans , mobilization , and escalating violence = =
& 11    & 952.6 & 370.4 & 60.2  & 32.3  & 55.8  & \textcolor{red}{1,690.1}   & \textcolor{red}{1,870.9}
\\\midrule
4 & = = = Loans to Peterborough and Molde = = =
& 11    & 295.6 & 110.5 & 31.3  & 19.9  & 22.7  & 216.6 & \textcolor{red}{488.2}
\\\midrule
5 & Gregzilla ( Greg Harrison ) — guitar ( 2015 – current )
& 12    & 711.1 & 519.5 & 117.5 & 39.5  & 41.5  & \textcolor{red}{50,441.8}  & \textcolor{red}{1,210.7}
\\\midrule
6  & 
Above the cut @-@ off frequency the image impedance is imaginary ,
& 12    & 1,075.6   & 505.3 & 103.7 & 58.3  & 57.5  & 500.1 & \textcolor{red}{1,131.9}     
\\\bottomrule
\end{tabularx}}
\label{tab:example_summary}
\end{table*}

In order to have an intuitive feeling, we selected data from the testing set as a case study. Table \ref{tab:example_unfairlen} and Table \ref{tab:example_summary} report the detail, and reveal that the PPL metric is unreliable.

Table \ref{tab:example_unfairlen} shows that long texts are more likely to get lower values. But when it is sliced into multiple sub-sentences, the PPL of each piece always rise substantially, even if they use the same word sequence as the original long texts.

As reported in Table \ref{tab:example_summary}, PPL is deeply perturbed by unusual text span duplication. It shows the trend that PPL decrease is greater as the length of the duplicated text increases (\textit{Orig.} --> $q1, k3$ --> $q1, k9$ --> $q5, k3$). This indicates that low-quality text can easily fool the metric by a large number of repetitions and achieve lower PPL values, making PPL metric less trustworthy.

PPL is also too sensitive to punctuation.
For example, after removing the last punctuation in the third sentence, the PPL rises to 196.4\% of the original text (increased from 952.6 to 1870.9).
And even the removal of punctuation of little significance (e.g. full stop) does not escape the disaster of PPL fluctuations. For example, in the first sentence, ppl drops to 76.1\% of the original sentence (decrease from 140.1 to 106.6) after removing the last mark.

Perturbations from unusual text span and punctuations are common problems in generative models. However, the above-mentioned cases tell us that these problems can easily break the PPL metric, cause large fluctuations and make it difficult to score the text quality fairly.

\end{document}